\documentclass{article}

\usepackage{arxiv}

\usepackage[utf8]{inputenc} 
\usepackage[T1]{fontenc}    
\usepackage{hyperref}       
\usepackage{url}            
\usepackage{booktabs}       
\usepackage{amsfonts}       
\usepackage{nicefrac}       
\usepackage{microtype}      
\usepackage{graphicx}
\usepackage{float}
\usepackage{threeparttable}
\usepackage{amsmath}
\usepackage[inkscapelatex=false]{svg}
\usepackage{geometry,color,graphicx,float}
\usepackage{tabularx}
\usepackage{svg}
\usepackage{algorithm} 
\usepackage{algpseudocode} 
\usepackage{adjustbox}
\usepackage{subcaption}
\usepackage{natbib}

\usepackage{doi}

\title{Synthetic Fungi Datasets: A Time-Aligned Approach}

\author{ \href{https://orcid.org/0000-0000-0000-0000}{\includegraphics[scale=0.06]{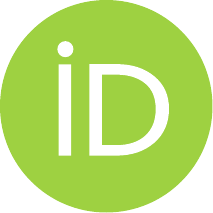}\hspace{1mm}Anju Rani}\thanks{Use footnote for providing further
		information about author (webpage, alternative
		address)---\emph{not} for acknowledging funding agencies.} \\
	Department of Energy Technology\\
	Aalborg University\\
	Esbjerg, Denmark 6700 \\
	\texttt{aran@energy.aau.dk} \\
	\And
	\href{https://orcid.org/0000-0000-0000-0000}{\includegraphics[scale=0.06]{orcid.pdf}\hspace{1mm}Daniel O.~Arroyo} \\
	Department of Energy Technology\\
	Aalborg University\\
	Esbjerg, Denmark 6700 \\
	\texttt{doa@energy.aau.dk} \\
	\And
	\href{https://orcid.org/0000-0000-0000-0000}{\includegraphics[scale=0.06]{orcid.pdf}\hspace{1mm}Petar Durdevic} \\
	Department of Energy Technology\\
	Aalborg University\\
	Esbjerg, Denmark 6700 \\
	\texttt{pdl@energy.aau.dk} \\
}

\renewcommand{\shorttitle}{Synthetic Fungi Datasets: A Time-Aligned Approach}

\hypersetup{
pdftitle={A template for the arxiv style},
pdfsubject={q-bio.NC, q-bio.QM},
pdfauthor={David S.~Hippocampus, Elias D.~Striatum},
pdfkeywords={First keyword, Second keyword, More},
}

\begin{document}
\maketitle

\begin{abstract}
Fungi undergo dynamic morphological transformations throughout their lifecycle, forming intricate networks as they transition from spores to mature mycelium structures. To support the study of these time-dependent processes, we present a synthetic, time-aligned image dataset that models key stages of fungal growth. This dataset systematically captures phenomena such as spore size reduction, branching dynamics, and the emergence of complex mycelium networks. The controlled generation process ensures temporal consistency, scalability, and structural alignment, addressing the limitations of real-world fungal datasets. Optimized for deep learning (DL) applications, this dataset facilitates the development of models for classifying growth stages, predicting fungal development, and analyzing morphological patterns over time. With applications spanning agriculture, medicine, and industrial mycology, this resource provides a robust foundation for automating fungal analysis, enhancing disease monitoring, and advancing fungal biology research through artificial intelligence.

An up-to-date GitHub repository accompanies this work, providing access to the dataset and supplementary materials: \href{https://github.com/PetarDurdevic/Funghi}{Synthetic Fungi Generation}
\end{abstract}


\section{Introduction}

In fungal biology, time-series alignment is essential for studying dynamic processes such as fungal growth, spore formation, and responses to environmental changes. Morphological changes in fungi are highly time-dependent and influenced by factors like nutrient availability, temperature fluctuations, and stress conditions. Aligning time-series imaging datasets ensures accurate comparisons of growth stages and morphological transitions across experiments or fungal strains, minimizing variability caused by external factors. This alignment enables the identification of consistent patterns and trends critical for understanding fungal behavior. Time-series datasets, particularly imaging-based ones, are invaluable for examining fungal dynamics, including responses to environmental or chemical stimuli. Such datasets have broad applications in medicine \cite{tahir2018fungus, picek2022automatic}, agriculture \cite{mahlein2016plant, de2015automated}, and industry \cite{de2019analysis}. For example, in agriculture, they support early detection of fungal infections on crops, improving disease management strategies. In medicine, they advance diagnostic tools and treatments for fungal infections. In industrial biotechnology, they optimize processes like enzyme production and bioremediation.

The time-aligned synthetic image dataset introduced in this paper is specifically designed to simulate realistic fungal growth scenarios, addressing challenges associated with inconsistent or limited real-world data. Fungi undergo dynamic morphological changes that require precise tracking across time to understand critical processes such as growth, spore formation, and responses to environmental stimuli. This dataset provides a systematic and controlled resource for studying these biological phenomena, enabling accurate modeling of fungal behavior under various conditions. In agriculture, time-aligned fungal datasets are invaluable for predicting fungal growth on crops, facilitating early interventions to mitigate crop losses by detecting infections before they spread. In medical research, these datasets support the study of pathogenic fungi, contributing to the development of more effective diagnostic and therapeutic strategies. By incorporating temporal data through techniques such as attention mechanisms or fusion networks, the dataset ensures that the time-dependent aspects of fungal growth are encoded effectively, enhancing tasks like spore detection and growth stage prediction. The integration of spatial and temporal information enables precise alignment with the complex dynamics of fungal behavior, making the dataset a powerful tool for applications in agriculture, medicine, and beyond. Additionally, advances in multimodal embedding techniques foster a deeper understanding of fungal processes, paving the way for scalable, high-precision tools to address real-world challenges across multiple domains.

These datasets are highly compatible with deep learning (DL) techniques, enabling a wide range of applications. DL models trained on the dataset can classify fungal species, identify growth stages, detect anomalies, and predict fungal behavior over time \cite{murugan2019estimation, tahir2018fungus, zielinski2020deep}. Such capabilities have significant implications for automating fungal monitoring systems, enhancing the efficiency of fungal identification in both laboratory and field environments, and advancing biotechnological processes where fungi play a key role, such as enzyme production and bioremediation. By leveraging synthetic datasets, researchers can address challenges like the scarcity of real-world data while developing scalable, high-precision tools to tackle complex problems involving fungi. Designed with expert analysis, this dataset maintains scientific rigor and faithfully represents the intricacies of fungal growth and behavior, making it a valuable resource for DL-driven research and applications.

\begin{figure}[h]
\centering
\includegraphics[width=8cm]{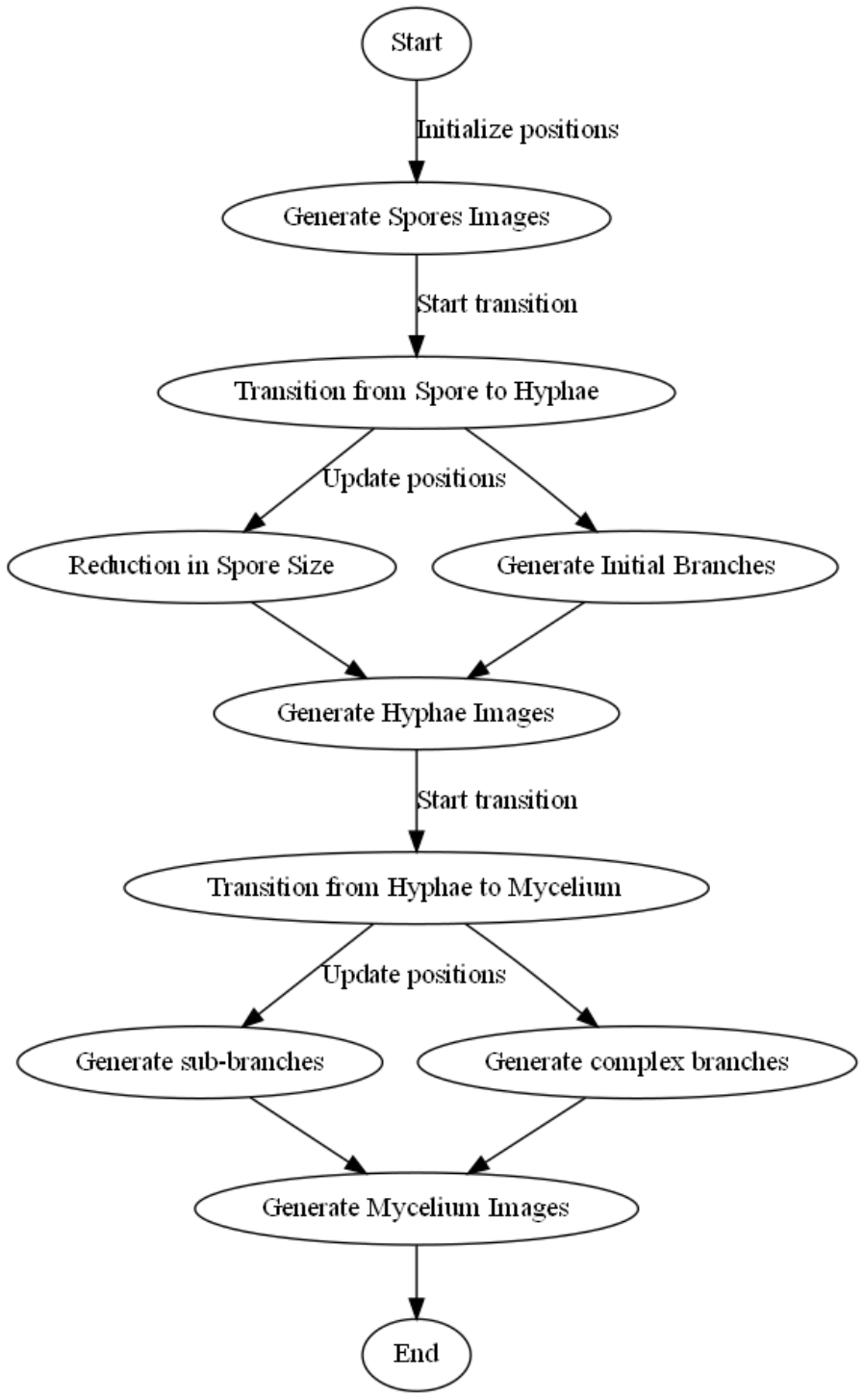}
\caption{Flowchart of the fungi generation process.}
\label{Fig1}
\end{figure}

\begin{table*}[ht]
\caption{Underlying mechanisms of the generated fungi dataset.}
\label{tbl1}
\centering
\begin{tabular}{p{0.17\linewidth}p{0.19\linewidth}p{0.24\linewidth}p{0.29\linewidth}}
\hline
\textbf{Feature} & \textbf{Probability Function} & \textbf{Parameters} & \textbf{Description} \\ 
\hline
Spore Positions & Uniform & Bounds: \([a_x, b_x]\), \([a_y, b_y]\) & Random initial spread of spores \\
Spore Jitter & Normal & \(\mu = 0, \sigma\) & Dynamic movement of spores over time \\
Branch Count & Poisson & \( \lambda \) & Natural variability in branching \\
Branch Length & Normal & \(\mu\), \(\sigma\) & Variability in growth size \\
Branch Angle & Uniform & \(\left[ 0, 2\pi \right]\) & Randomized directionality of branches \\
Phase Transition & Quadratic & \(T\) & Smooth transitions between growth stages \\
Growth Factor & Quadratic Growth & \(T\), \(\tau\) & Simulates progression and acceleration \\
Temperature Factor & Normal & \(\mu = 1, \sigma_T\) & Environmental variability affecting growth \\
Mycelium Density & Exponential Decay & \(\alpha\) & Models interconnected web-like structures \\
\hline
\end{tabular}
\end{table*}

\begin{figure*}
\centering
\includegraphics[width=\textwidth]{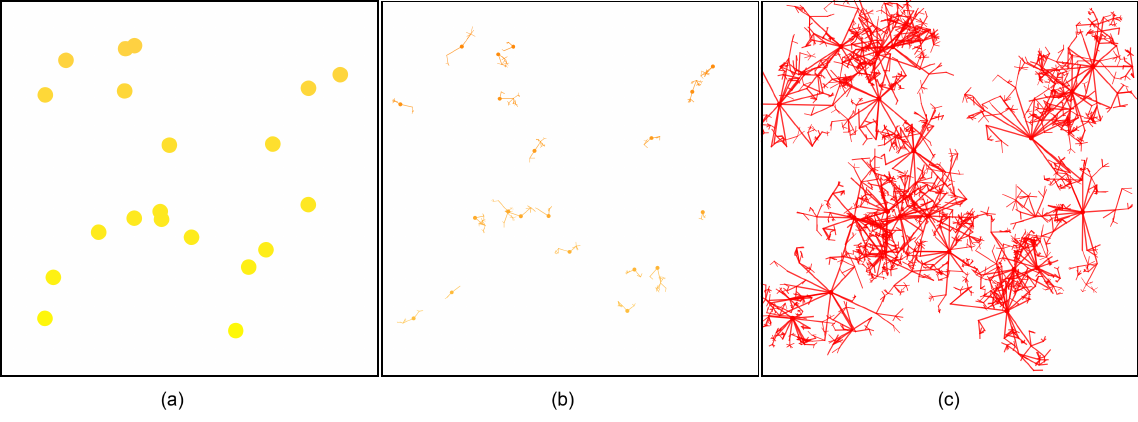}
\caption{Synthetic generated fungi dataset: (a) Spore, (b) Hyphae, and (c) Mycelium.}
\label{Fig2}
\end{figure*}

\begin{figure*}
\centering
\includegraphics[width=\textwidth]{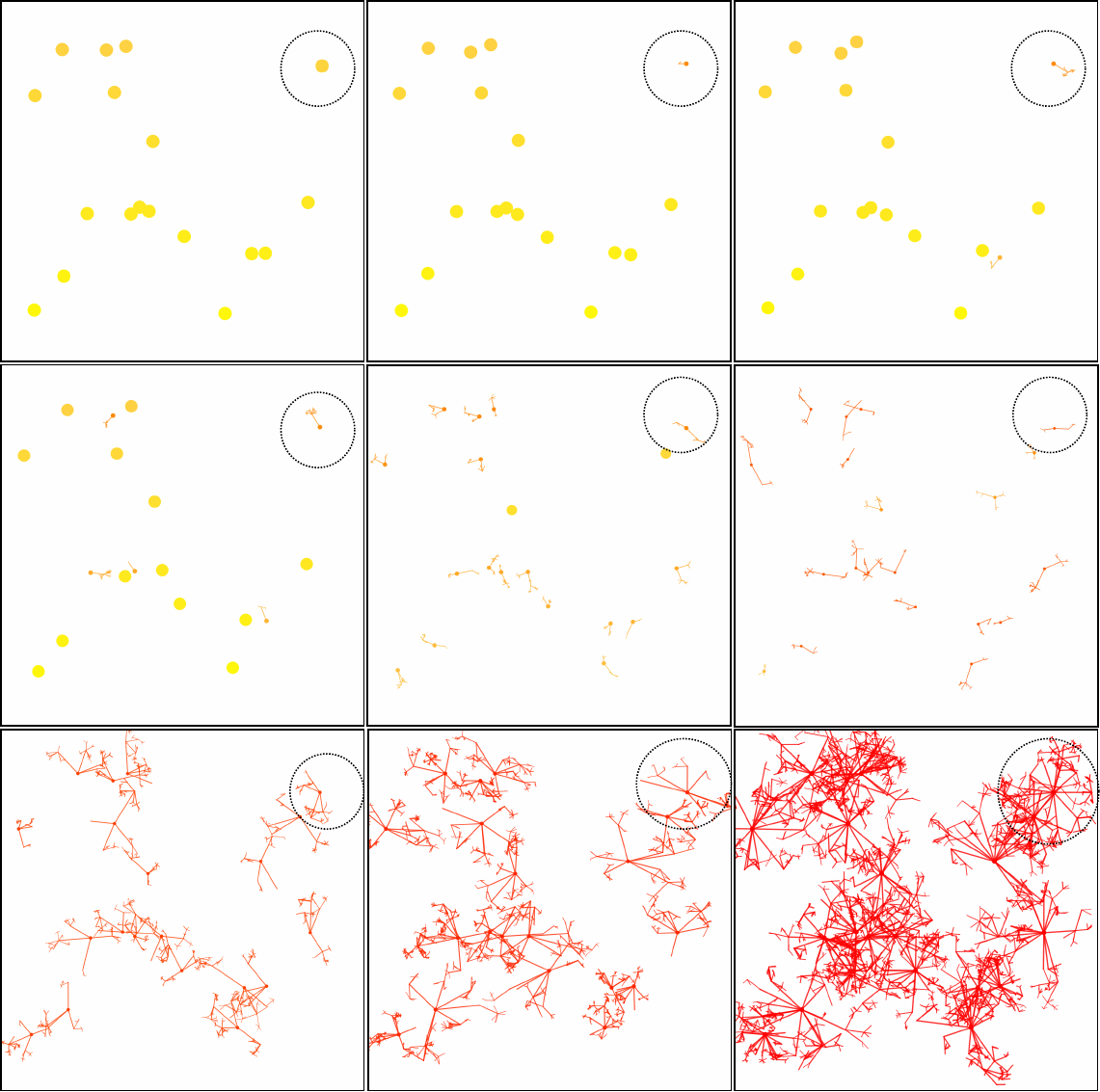}
\caption{Real-time tracking of the generated fungi dataset.}
\label{Fig3}
\end{figure*}

The structure of the paper is as follows: Section 2 describes the methodology employed in this study, providing an overview of the process used to generate a synthetic fungi dataset. Section 3 discusses potential future work related to the generated dataset. Lastly, Section 4 presents the conclusions, summarizing our findings and reflecting on their implications.

\section{Generation of Fungi Dataset}
The synthetic fungi dataset is designed to replicate the growth of fungi, progressing through stages such as spores, hyphae, and mycelium. It generates a series of images that visually depict these biological processes, with several innovative features for customization and randomness. Figure~\ref{Fig1} depicts the flowchart of the fungi generation process involved in this paper. This paper highlights the concept of recursive branching to accurately simulate the growth dynamics of fungal structures like hyphae and mycelium. In the dataset, recursive branching is implemented with probabilistic controls to ensure a natural appearance. The growth of branches is modeled by adjusting branch length and width according to the depth in the recursive structure. This scaling is further influenced by growth factors and random variability, utilizing normal and uniform distributions. The number of sub-branches emerging from a node follows a Poisson distribution, capturing the natural variability in fungal growth.

To improve realism, sub-branch orientations are assigned random angles uniformly distributed between \(0\) and \(2\pi\), promoting diverse, multidirectional growth patterns. Environmental factors, such as temperature, are also incorporated into the model. These temperature factors influence both branch length and width, introducing variability based on external conditions. This approach mirrors the effects of environmental stresses or stimuli, which are known to significantly impact fungal morphology and growth dynamics.

\begin{algorithm}
\fontsize{10}{12}\selectfont 
    \caption{Generate Fungi Transition Images}
    \begin{algorithmic}
        \State Initialize parameters: $num\_images$, $growth\_factors$, $size\_factors$.
        \State Initialize lists for $spores$, $hyphae$, $mycelium$.

        \Function{InterpolateColor}{$start\_color, end\_color, position$}
            \State Return interpolated color.
        \EndFunction

        \Function{GenerateBranch}{$x, y, branch\_length, angle, depth, width, color$}
            \If {depth = 0 or $branch\_length < threshold$} \Return \EndIf
            \State Compute $end\_x, end\_y$ and draw branch.
            \ForAll {sub-branches}
                \State Sample $new\_angle, new\_branch\_length, new\_branch\_width$.
                \State \Call{GenerateBranch}{}.
            \EndFor
        \EndFunction

        \Function{VisualSpores}{$spores, num\_spores, size\_factor$}
            \ForAll {$spores$} 
                \State Draw spore with computed size and color.
            \EndFor
        \EndFunction

        \Function{VisualHyphae}{$hyphae, num\_hyphae, size\_factor, growth\_factor$}
            \ForAll {$hyphae$}
                \State Draw hypha with computed size and color.
            \EndFor
        \EndFunction

        \Function{VisualMycelium}{$mycelium, num\_mycelium, growth\_factor$}
            \ForAll {$mycelium$}
                \State Draw mycelium with computed size and color.
            \EndFor
        \EndFunction

        \For {$i = 1$ to $num\_images$}
            \State $transition\_ratio \gets i / num\_images$.
            \State Adjust $num\_spores$, $num\_hyphae$, $num\_mycelium$.
            \State Convert $spores \to hyphae$, $hyphae \to mycelium$.
            \State \Call{VisualSpores}{}, \Call{VisualHyphae}{}, \Call{VisualMycelium}{}.
            \State Save image.
        \EndFor
    \end{algorithmic}
\end{algorithm}

\subsection{Dynamics of Fungi Generation}
The generation of spores, hyphae, and mycelium is influenced by a blend of spatial positioning, temporal transitions, and stochastic processes. Table \ref{tbl1} presents the underlying mechanisms of the generated fungi dataset.

\subsubsection{Spore Dynamics}
\textbf{Spatial Aspect (Positioning of Spores)}: The spores are initially placed at random positions within a confined 2D space, ensuring that the starting positions are centered within the plot. Each coordinate (x, y) is independently drawn from a uniform distribution within the range [0.1, 0.9]. At each iteration, the spores undergo slight random movement to simulate diffusion or environmental fluctuations. The updated positions are calculated as follows:

\begin{equation}
\left.\begin{matrix}
x_{new} = x+\delta_x \\
y_{new} = y+\delta_y  \\ 
\end{matrix}
\right\} \delta_x, \delta_y \sim U(-0.01, 0.01)
\end{equation}

where \((x,y)\) represents the cartesian coordinates of a spore, hypha, or mycelium in 2D space and \((x_{new}, y_{new})\) are the new coordinates of a branch's endpoint.

\textbf{Temporal Aspect (Evolution of Spores)}: The temporal aspect governs the transformation of spores into hyphae and mycelium. Over time, the size of each spore gradually decreases, reflecting its depletion as it transitions into hyphae. This shrinking process is controlled by the transition ratio \(T\), which is determined based on the index of the current image, calculated as:

\begin{equation}
\text{T} = \frac{i}{N - 1}, \quad i \in [0, N - 1]
\end{equation}

where, \(i\) is the current iteration (or frame) index, and \(N\) is the total number of frames. As the \(T\) increases, the number of spores gradually transforms into hyphae, as described by:

\begin{equation}
S(t)=S_0\ast (1-T)
\end{equation}

where \(S_0\) is the initial spore count, and \(S(t)\) is the number of spores at time \(t\). A fraction of spores transitions into hyphae, with the number of converted spores directly proportional to the \(T\) (e.g., \(T\) = 0.1), expressed as \(H(t)\propto T*S(t)\). Similarly, hyphae begin to form mycelium once \(T\) exceeds a predefined threshold (e.g., \(T\) = 0.5), as represented by \(M(t)\propto max(0, T-0.5)\).

\subsubsection{Hyphae Dynamics}
\textbf{Spatial Aspect}: Hyphae, the mid-stage of fungal growth, emerge from spores. Their initial positions are inherited from the spore locations that transition into hyphae. As new branches grow, their spatial positions extend outward from the parent location. Hyphae exhibit a fractal-like growth pattern, with each segment branching into multiple sub-branches. The spatial extension of each branch is determined by its length \(L\) and angle \(\theta\):

\begin{equation}
\begin{split}
x_{new} & = x_{current} + L* cos(\theta ) \\
y_{new} & = y_{current} + L* sin(\theta )
\end{split}
\end{equation}

\textbf{Temporal Aspect}: The recursive depth of branching governs the temporal growth of hyphae. With each iteration, the branches decrease in both length \((L)\) and width \((W)\), reflecting real-world energy limitations:
\begin{equation}
\begin{split}
L_{branch} & = L_{previous}*Decay\_Factor \\
W_{branch} & = W_{previous}*Decay\_Factor
\end{split}
\end{equation}

Similar to the spores, some hyphae gradually transition into mycelium over time, driven by the \(T\). 

\textbf{Probabilistic Aspect}: Stochastic processes are integral to the branching dynamics and the random behavior of all three stages of fungal growth. The transition from spores to hyphae involves selecting a random subset of spores for conversion into the hyphae stage. While the number of spores to be converted is determined deterministically by the \(T\) the actual selection of spores is applied stochastically, with spores being randomly chosen for the transition.

\emph{Branch count}: The branching dynamics are modeled probabilistically, with the number of sub-branches following a Poisson distribution, given by:
\begin{equation}
n_{sub-branches} \sim Poisson(\lambda)
\end{equation}

where \(\lambda\) represents the average number of sub-branches.

\emph{Branch length \((L)\)}: To imitate the natural growth of fungi, a random factor drawn from a normal distribution is used to scale each branch length:
\begin{equation}
L_{branch} \sim \mathbb{N}(\mu, \sigma)
\end{equation}

where \(\mu\) and \(\sigma\) represent the mean length and standard deviation, respectively.

\emph{Branch angle \((\theta)\)}: Angles are sampled uniformly across all directions, as given by: \(\theta \sim U(0, 2\pi)\).

\emph{Branch Width \((W)\)}: The widths of branches are scaled by a uniform factor, given by: \(W_{branch} \sim U(0.6, 1.0)\).

Furthermore, the evolution of a branch follows a recursive process, influenced by its depth and parameters such as length and width. For each branch, the next branch length is determined by:
\begin{equation}
L_{next} = L_{current}*0.7*\mathbb{N}(1, 0.2)
\end{equation}

This recursion halts when the depth reaches zero or the branch length becomes negligibly small.

\subsubsection{Mycelium Dynamics}

\textbf{Spatial Aspect}: Similar to hyphae, the positions of mycelium are inherited from the hyphae that transition into mycelium. The final stage of fungal growth, mycelium, forms a dense network by expanding outward from the initial position. Figure~\ref{Fig2} represents the synthetically generated fungi dataset. The spatial growth of mycelium is influenced by dynamic branch lengths and random angles, similar to the growth of hyphae. However, mycelium forms a more intricate network compared to hyphae, as it undergoes increased branching and exhibits less decay in branch length.

\textbf{Temporal Aspect}: The growth factor of mycelium increases with \(T\), as expressed by \(Growth\_Factor \propto max(0, T-0.5)\). Mycelium branches are more persistent than hyphae, exhibiting slower decay rates for both branch length and width. This slower decay reflects the role of mycelium in establishing a stable and enduring fungal network.

\textbf{Probabilistic Aspect}: Mycelium branches, which have more sub-branches per node, follow a stochastic model similar to hyphae, but with a higher branching density. The branch lengths and widths of mycelium are sampled using slightly different parameters to reflect the denser and thicker structure of mycelium, as given by:
\begin{equation}
\begin{split}
L \sim \mathbb{N}(\mu = 0.8, \sigma = 0.15) \\
W \sim U(0.7, 1.0)
\end{split}
\end{equation}

Figure~\ref{Fig3} illustrates the real-time tracking of the generated fungi dataset, showcasing the dynamic progression of fungal growth as it evolves over time. Figure~\ref{Fig3} provides a visual representation of the transition from individual spores to branching hyphae and ultimately to the formation of a dense mycelium network. Each frame in the real-time tracking highlights the spatial and temporal changes in the fungal structures, capturing the natural variability and complexity of growth patterns. This visualization serves as a valuable tool for monitoring the evolution of fungal growth, allowing researchers to observe and analyze the growth dynamics in real time. Additionally, the tracking system can be used to validate the accuracy of the synthetic dataset and to ensure the alignment of growth stages throughout the simulation. The ability to monitor these processes in real-time also aids in the development of deep learning models for automated fungal analysis and classification. Thus, the model captures the structural complexity and natural variability observed in fungal growth systems.

\section{Future Work}
This section explores how the synthetic dataset can serve as a foundation for studying fungal systems in various contexts, thereby enhancing the understanding of these crucial organisms in both natural and engineered environments.

\begin{enumerate} 
\item \textbf{Training Advanced Computer Vision Models}: The synthetic dataset can be used to train deep learning (DL) models for fungal growth recognition, segmentation, and classification. This is particularly valuable for automating fungal identification in agricultural or ecological research. 
\item \textbf{Simulating Complex Ecosystems}: The dataset offers a platform for simulating interactions between fungi and other environmental factors, making it ideal for modeling soil ecosystems or forest networks in virtual settings. 
\item \textbf{Synthetic-to-Real Domain Adaptation}: Researchers can apply domain adaptation techniques to bridge the gap between synthetic fungal images and real-world microscopy data, improving the generalization of AI models trained on this dataset. 
\item \textbf{Algorithm Benchmarking}: The dataset can be used as a standardized benchmark to evaluate algorithms in fungal growth modeling, including simulation accuracy, growth rate prediction, and network structure analysis. 
\item \textbf{Exploration of Structural Patterns}: Researchers can use the dataset to study how various environmental and probabilistic parameters influence fungal structures, which could lead to hypotheses and tests of biological phenomena. 
\item \textbf{Bio-inspired Network Design}: Engineers could leverage the dataset to explore fungal network patterns for bio-inspired applications in network optimization, resource distribution, or robotics. 
\item \textbf{Synthetic Biology Applications}: Synthetic biologists may use the dataset to design and test hypotheses for fungal growth control or manipulation, enabling advancements in biotechnology applications. 
\end{enumerate}

\section{Conclusion}
The time-aligned fungi dataset provides a robust synthetic framework for simulating the dynamic growth of fungi, encompassing the transitions from spores to hyphae and ultimately to dense mycelium networks. By integrating spatial, temporal, and probabilistic factors, the dataset captures essential morphological features and growth patterns, offering a scalable and controlled resource to address challenges associated with real-world data limitations. Its design ensures temporal alignment, supporting the development of deep learning models for fungal growth recognition, classification, and prediction.

This dataset holds significant potential for applications in agriculture, medicine, and industrial mycology. It can be used for automated monitoring of fungal infections in crops, the study of pathogenic fungi in medical research, and optimization processes in biotechnology. Additionally, the dataset enables researchers to explore structural patterns, benchmark algorithms, and train models for real-world fungal analysis. The use of synthetic data bridges gaps in experimental data availability and improves the understanding of fungal systems under various environmental conditions.

Future work could expand the dataset's capabilities by incorporating interactions with environmental stimuli, enabling 3D modeling for more realistic ecological simulations, and applying domain adaptation techniques to integrate synthetic data with real-world microscopy images. This dataset not only provides a foundation for advancing fungal biology research but also opens doors for interdisciplinary applications, such as bio-inspired network design and innovations in synthetic biology.

\section*{Ethics Statement}
This research does not involve experiments, observations, or data collection related to human or animal subjects. 

\section*{Declaration of competing interest}
The authors declare that they have no known competing financial interests or personal relationships that could have appeared to influence the work reported in this paper.

\section*{Data Availability}
\href{https://github.com/PetarDurdevic/Funghi} {Synthetic Fungi Generation}. 

\bibliographystyle{unsrtnat}
\bibliography{references}  
\end{document}